# Efficient Evaluation of the Number of False Alarm Criterion

**Sylvie Le Hégarat-Mascle · Emanuel Aldea · Jennifer Vandoni**

**Abstract** This paper proposes a method for computing efficiently the significance of a parametric pattern inside a binary image. On the one hand, a-contrario strategies avoid the user involvement for tuning detection thresholds, and allow one to account fairly for different pattern sizes. On the other hand, a-contrario criteria become intractable when the pattern complexity in terms of parametrization increases. In this work, we introduce a strategy which relies on the use of a cumulative space of reduced dimensionality, derived from the coupling of a classic (Hough) cumulative space with an integral histogram trick. This space allows us to store partial computations which are required by the a-contrario criterion, and to evaluate the significance with a lower computational cost than by following a straightforward approach. The method is illustrated on synthetic examples on patterns with various parametrizations up to five dimensions. In order to demonstrate how to apply this generic concept in a real scenario, we consider a difficult crack detection task in still images, which has been addressed in the literature with various local and global detection strategies. We model cracks as bounded segments, detected by the proposed a-contrario criterion, which allow us to introduce additional spatial constraints based on their relative alignment. On this application, the proposed strategy yields state-of the-art results, and underlines its potential for handling complex pattern detection tasks.

Sylvie Le Hégarat-Mascle, Emanuel Aldea, Jennifer Vandoni
SATIE Laboratory, Building 660, rue Noetzlin, Université Paris Sud, Orsay, France
E-mail: sylvie.le-hegarat@u-psud.fr

## 1 Introduction

Since the seminal articles of Desolneux et al. [10,11], detection approaches based on the Number of False Alarms criterion (NFA) became more and more popular in the field of image processing this last decade. In these approaches, the words 'a-contrario' refer to the fact that detection is performed by contradicting a 'naive' model that represents the statistics of the outliers (the null hypothesis in statistical decision theory). Then, the inliers are detected as too regular to appear 'by chance' according to the naive model. Defining a measurement of deviation relatively to the naive model, the main asset of such approaches is their independence from threshold parameters, as they cast the detection as an optimization problem by maximizing the *significance* defined from deviation. Then, to interpret this maximum of significance (or minimum NFA value) in terms of presence or absence of structured pattern, one refers to the NFA definition itself: the NFA of a pattern candidate is the expected number of false positives (random patterns) occurring in the search space when accepting all patterns at least as significant as the candidate. By setting the NFA detection threshold to 1 irrespective of the detection task, the a-contrario framework simply states that the upper bound for detecting a random rare event is at most one occurrence.

Since the introductory illustration of a-contrario methods on alignment detection [10], some recent works developed the underlying idea [18,7], but a large interest developed around using this fundamental pattern grounded in the Gestalt continuity principle in order to detect related elements such as segments [29,30,1], vanishing points [4,17] or scratches [19]. Concomitantly, a-contrario methods have been developed for detecting more complex patterns, such as circles and ellipses [2,22] as well as coherent clusterings in a broader sense [28,8,25,26,21,32].

Considering pattern recognition problems, in order to find the most significant subset among the ones representing patterns, one should theoretically compute the *significance* of every possible pattern. If the researched patterns correspond to parametric objects (e.g. lines, ellipses etc.), the dimension (and thus the actual cardinality being used) of the solution space to explore grows with the number of parameters. Then, in order to maintain tractability, discretizing the parameter space or relying on heuristic strategies [30] for exploration are commonly employed. In this work, we show how a cumulative space may be used in order to compute efficiently the significance of a given parametric pattern. Cumulative approaches, widely used in pattern recognition and introduced in [15], rely on a quantization of the entire feasible parameter space denoted as accumulator, in which every observation increments the count of every discrete cell (i.e pattern) consistent with the existence of that observation. In the end, each cell records the total number of supporting observations.

In our work we show that in some favorable cases there is an equivalence between considering cells in a high dimensionality cumulative space and recasting them as n-orthotopes in an alternative cumulative space of decreased dimensionality (by 1 or 2 in the proposed examples). Now, using a trick similar to the integral histogram [23], the fact of considering n-orthotopes allows for an efficient computation of the required NFA values. Specifically, the integral histogram is the result of the propagation of an aggregated histogram from origin through the whole image lattice so that the histogram of any rectangular region may be computed by simple arithmetic operations between four points of the integral histogram. Thus, privileging this cumulative space of reduced dimensionality allows us to lower significantly the complexity of pattern research, and to extend the limits of the parameter space dimension due to the available computer memory size.

Then, as a second contribution, we show how crack detection in still images can benefit from the proposed coupling between an a-contrario criterion and a cumulative space.

## 2 NFA computation using a cumulative space

### 2.1 Related work

In previous works involving NFA criterion, two 'naive' models have been widely used, namely the Gaussian and Bernoulli models that stand for gray level and binary images, respectively. In this study, we focus on the second case. Pixels take then values in $\{0, 1\}$ set (or $\{false, true\}$ set). Assuming a Bernoulli distribution of parameter $p$ for pixel binary values, the probability to have a given number $\kappa$ of $true$ samples, i.e. 1-valued pixels, among a given number $\nu$ of pixels is a Binomial distribution of parameter $p$. Then, according to [13], the Number of False Alarms is

$$NFA_B\left(\kappa, \nu, p\right) = \eta_2 \sum_{i=\kappa}^{\nu} \binom{\nu}{i} p^i \left(1 - p\right)^{\nu - i},$$  (1)

where $\eta_2$ is the 'number of tests' coefficient that depends on the number of possible patterns of $\nu$ pixels.

The significance being defined as $S\left(\kappa, \nu, p\right) = -\ln(NFA_B(\kappa, \nu, p))$, it may be derived from Eq. (1). Using the Hoeffding's approximation like in [12], we may express $S\left(\kappa, \nu, p\right)$ in terms of the Kullback-Leibler divergence (K-L) between the $true$ pixel probability restricted to a particular area, $p_a = \frac{\kappa}{\nu}$, and the naive model probability $p$: $\forall \left(\kappa, \nu\right)$ such that $\frac{\kappa}{\nu} > p$,

$$S\left(\kappa, \nu, p\right) \approx \nu \left[\frac{\kappa}{\nu} \ln\left(\frac{\kappa/\nu}{p}\right) + \left(1 - \frac{\kappa}{\nu}\right) \ln\left(\frac{1 - \kappa/\nu}{1 - p}\right)\right] - \ln \eta_2$$  (2)

## 2.2 Proposed approach

According to Eq. (1) or Eq. (2), in order to compute the *significance* of a given pattern, we need both its geometric area or its number of pixels and its number of *true* pixels.

The main idea of this work is to use a cumulative space to store partial sums of numbers of points in order to decrease the computational cost. Then, the number of points in any pattern of given parameters can be directly retrieved from the values stored in the cumulative space, allowing us to accelerate the algorithm and/or to cope with a finer discretization of the parameter space.

In the following part of this section, we explain our approach and we illustrate its operation through four classic examples, namely detection of rectangular tiles, strips, rings and bounded strips.

### 2.2.1 Use of cumulative space

The cumulative space on which we will focus varies with the considered pattern. Specifically, it derives from the chosen parametric form for the pattern of interest. Now, all the parametric forms (of a given pattern) are not equivalent in terms of involved cumulative space. Let us point out the representations as a set of 'simpler' patterns (simpler in the sense that they involve less parameters), such that the set is defined by varying one (or two) parameter(s) of the simpler pattern into an interval. For instance, a strip can be represented either as a straight line having a strictly positive width or as a set of parallel lines such that their distance to the origin ($\rho$ parameter in polar representation) varies between two bounds. Now, let us remark that, for such a parametric form, when two parameters represent the bounds of an interval, it is possible to handle them on a single axis/dimension of the associated cumulative space. In the example of the strip, the first representation involves a 3D cumulative space, whereas the second representation allows us to use the same 2D cumulative space as that of straight lines, namely the classic Hough transform space.

Therefore, among several parametric forms of a given pattern, denoted $\mathbf{b}$, we favor the one that is a set of simpler patterns denoted $\mathbf{a}$, having several parameters that we will be able to represent on a same axis of the cumulative space. Such a representation allows us to reduce the dimensionality of the cumulative space and save processing time by storing partial sums, in a similar fashion to integral histograms [23]. It allows us to compute the pattern as follows.

Let $l$ denote the number of parameters required to determine the considered pattern $\mathbf{b}$. We denote by $\boldsymbol{\beta} = (\beta_i)_{i \in [1,l]}$ the tuple of these parameters which take values in $\mathcal{A}$ ($\mathcal{A}$ depends on the image lattice and on the application that may introduce some specific constraints on the parameters).

Let $\mathcal{C}$ be the considered cumulative space. If $\mathbf{b}$ pattern has been parametrized as a set of 'well-chosen' $\mathbf{a}$ patterns, $\mathcal{C}$ is the cumulative space associated to **a** parametrization. Since **a** has less parameters than **b**, some pairs of **b** parameters are bounds for intervals of **a** parameters. Then, we distinguish in $\mathcal{C}$ the axes that represent only one parameter $\beta_i$, and the axes that represent two different $\beta_i$ (playing the role of bounds for some **a** parameters). If $m$ is the number of bi-parameter axes, with $0 \leq m \leq \frac{l}{2}$, $\mathcal{C}$ dimensionality is $l - m$, and $l - 2m$ is the number of mono-parameter axes. Note that in $\mathcal{C}$ a simpler pattern is represented by a point and a pattern of interest by a n-orthotope (also called hyperrectangle). Indeed, any pattern of interest **b** is then represented by a n-orthotope of $\mathcal{C}$ having $l - 2m$ dimensions reduced to a single point and the other dimensions which are non-null intervals. Now, since $\mathcal{C}$ is a cumulative space associated to **a** pattern, the number of votes for a given **a** pattern is provided by the value of the corresponding point in $\mathcal{C}$ and the number of votes for a given **b** pattern is the sum of the $\mathcal{C}$ point values over the corresponding hypercube.

Finally, following the integral histogram idea [23], we have to compute and store partial sums over $\mathcal{C}$. To be able to specify the partial sum computation, we have to order the parameters.

Without loss of generality, the elements of $\boldsymbol{\beta}$ tuple are mapped to the $\mathcal{C}$ axes, denoted $(\alpha_i)_{i \in [1, l-m]}$, as follows: the $l - 2m$ first components are mapped to the $l - 2m$ first $\mathcal{C}$ parameter axes (that are thus mono-parameter axes) and the $2m$ last components are mapped to the $m$ last $\mathcal{C}$ parameter axes (that are thus bi-parameter axes) so that the $\beta_i$ parameters are reordered as $\left((\alpha_i)_{i \in [1, l-2m]}, (\underline{\alpha}_j)_{j \in [l-2m+1, l-m]}, (\overline{\alpha}_j)_{j \in [l-2m+1, l-m]}\right)$ where $\underline{\alpha}_j$ denotes an interval lower bound and $\overline{\alpha}_j$ denotes an interval upper bound.

In the following, for conciseness and since we focus on bi-parameter axes, the list of parameters of mono-parameter axes (that may possibly be empty) is denoted by *dots*. Then, if $J_{\mathcal{C}}$ denotes the cumulative space instance where the value of a point represents its number of votes and if we choose a unit step for the cumulative space resolution, the cumulative space instance containing partial sums $\mathbb{J}_{\mathcal{C}}$ is computed as follows: if $m = 1$:

$$\mathbb{J}_{\mathcal{C}}\left(\ldots, \alpha_{l-1}\right) = J_{\mathcal{C}}\left(\ldots, \alpha_{l-1}\right) + \mathbb{1}_{[\alpha_{l-1} > 1]} \mathbb{J}_{\mathcal{C}}\left(\ldots, \alpha_{l-1} - 1\right), \quad (3)$$

where $\mathbb{1}_{[\alpha_i > 1]}$ denotes the indicator function that takes value 1 if the condition between square brackets is true and value 0 otherwise; and if $m = 2$,

$$\mathbb{J}_{\mathcal{C}}\left(\ldots, \alpha_{l-3}, \alpha_{l-2}\right) = J_{\mathcal{C}}\left(\ldots, \alpha_{l-3}, \alpha_{l-2}\right) - \mathbb{1}_{\left[\substack{\alpha_{l-3} > 1 \\ \alpha_{l-2} > 1}\right]} \mathbb{J}_{\mathcal{C}}\left(\ldots, \alpha_{l-3} - 1, \alpha_{l-2} - 1\right)$$
$$+ \mathbb{1}_{[\alpha_{l-3} > 1]} \mathbb{J}_{\mathcal{C}}\left(\ldots, \alpha_{l-3} - 1, \alpha_{l-2}\right) \qquad (4)$$
$$+ \mathbb{1}_{[\alpha_{l-2} > 1]} \mathbb{J}_{\mathcal{C}}\left(\ldots, \alpha_{l-3}, \alpha_{l-2} - 1\right)$$

From $\mathbb{J}_{\mathcal{C}}$, the number of *true* pixels in a given pattern such that $(\alpha_i, i \in [1, l-2m]) = \mathbf{b}_1$ (possibly nonexistent if $l = 2m$) and $\forall i \in [l - 2m + 1, l - m]$, $\alpha_i \in [\underline{\alpha}_i, \overline{\alpha}_i]$ is:

— if $m = 1$,

$$\kappa\left(\mathbf{b}\right) = \sum_{\alpha_{l-1} = \underline{\alpha}_{l-1}}^{\overline{\alpha}_{l-1}} J_{\mathcal{C}}\left(\mathbf{b}_1, \alpha_{l-1}\right) = \mathbb{J}_{\mathcal{C}}\left(\mathbf{b}_1, \overline{\alpha}_{l-1}\right) - \mathbb{J}_{\mathcal{C}}\left(\mathbf{b}_1, \underline{\alpha}_{l-1}\right), \quad (5)$$

– if $m = 2$,

$$
\begin{aligned}
\kappa\left(\mathbf{b}\right) &= \sum_{\alpha_{l-3}=\underline{\alpha}_{l-3}}^{\overline{\alpha}_{l-3}} \sum_{\alpha_{l-2}=\underline{\alpha}_{l-2}}^{\overline{\alpha}_{l-2}} J_{\mathcal{C}}\left(\mathbf{b}_{1}, \alpha_{l-3}, \alpha_{l-2}\right), \\
&= \mathbb{J}_{\mathcal{C}}\left(\mathbf{b}_{1}, \overline{\alpha}_{l-3}, \overline{\alpha}_{l-2}\right) + \mathbb{J}_{\mathcal{C}}\left(\mathbf{b}_{1}, \underline{\alpha}_{l-3}-1, \underline{\alpha}_{l-2}-1\right) \\
&\quad - \mathbb{J}_{\mathcal{C}}\left(\mathbf{b}_{1}, \underline{\alpha}_{l-3}-1, \overline{\alpha}_{l-2}\right) - \mathbb{J}_{\mathcal{C}}\left(\mathbf{b}_{1}, \overline{\alpha}_{l-3}, \underline{\alpha}_{l-2}-1\right).
\end{aligned}
\tag{6}
$$

In summary, the main idea of the paper is that the NFA can be computed more efficiently with cumulative space pre-computation: for an image of size $N^2$, using the integral histogram trick in a well-chosen cumulative space of reduced dimension $l - m$ (instead of $l$) with each single dimension of size $M$, the complexity is roughly $N^2$ ($J_{\mathcal{C}}$ computation) $+ M^m$ ($\mathbb{J}_{\mathcal{C}}$ computation) $+ M^l$. Then, reducing the complexity compared to a complete brute force approach in $N^2 \times M^l$ allows for more precise results by providing finer estimation of pattern parameters.

Let us now consider four classic examples, namely rectangular tiles, strips, rings and bounded strips, by specifying for each case the corresponding cumulative space. These patterns are detected on an image having $N_c$ columns and $N_r$ rows and half diagonal length $\rho_d = \frac{1}{2}\sqrt{N_c^2 + N_r^2}$.

– <u>rectangular tiles</u> are parametrized as sets of 2D points, using the coordinates of two opposite corners: $(x_{UL}, y_{UL}, x_{LR}, y_{LR})$ where $x_{UL}$ and $y_{UL}$ (respectively $x_{LR}$ and $y_{LR}$) denote the image coordinates (column and row) of the upper left (respectively lower right) corner of the tile; $x_{UL} \in [1, N_c]$, $y_{UL} \in [1, N_r]$, $x_{LR} \in [x_{UL}, N_c]$ and $y_{LR} \in [y_{UL}, N_r]$. The cumulative space $\mathcal{T}$ has two dimensions: the column axis $x$ representing $x_{UL}$ and $x_{LR}$ and the row axis $y$ representing $y_{UL}$ and $y_{LR}$. $J_{\mathcal{T}}$ is the binary image itself and $\mathbb{J}_{\mathcal{T}}$, the cumulative space containing $J_{\mathcal{T}}$ partial sums, is derived from Eq. (4) with $l = 4$, $m = 2$, $\alpha_{l-3} = x$ and $\alpha_{l-2} = y$.

– <u>strips</u> are parametrized as sets of parallel straight lines, through the polar coordinates of the two border lines: $(\theta, \rho_0)$ and $(\theta, \rho_1)$ where $\theta$ is the parallel line direction (one single value) and $\rho_0$ and $\rho_1$ the distance of the border lines to the space origin. Choosing the image center as origin, $\rho_0$ and $\rho_1$ are signed values so that there is no discontinuity in $\rho$ values for strips containing the origin. Then, $\theta \in [0, \pi)$, $\rho_0 \in [-\rho_d, \rho_d]$, $\rho_1 \in [\rho_0, \rho_d]$. The cumulative space $\mathcal{S}$ has two dimensions: the angular axis $\theta$ and the distance axis $\rho$ representing $\rho_0$ and $\rho_1$. $J_{\mathcal{S}}$ is the classic Hough transform [15], and $\mathbb{J}_{\mathcal{S}}$ is the cumulative space containing $J_{\mathcal{S}}$ partial sums, derived from Eq. (4) with $l = 3$, $m = 1$ and $\alpha_{l-1} = \rho$.

– <u>rings</u> are parametrized as sets of concentric circles, through the circle center coordinates $(x_0, y_0)$ and two rays $\rho_0$ and $\rho_1$ respectively, where $x_0 \in [1, N_c]$, $y_0 \in [1, N_r]$, $\rho_0 \in [0, \rho_d]$, $\rho_0 \in [\rho_0, \rho_d]$. The considered cumulative space $\mathcal{R}$ has three dimensions: the column and row axes $x$ and $y$ for the coordinates of the center and the ray axis $\rho$ representing $\rho_0$ and $\rho_1$. $J_{\mathcal{R}}$ is the circle Hough transform, and $\mathbb{J}_{\mathcal{R}}$ is the cumulative space containing $J_{\mathcal{R}}$ partial sums, derived from Eq. (4) with $l = 4$, $m = 1$ and $\alpha_{l-1} = \rho$.

– **bounded strips** are sets of parallel segment lines that can also be represented as unbounded strips with two extremities. They are parametrized by a 5-tuple $(\theta, \rho_0, \phi, \rho_1, \psi)$ where $(\theta, \rho_0, \rho_1)$ represents the unbounded strip as previously stated, and $\phi$, $\psi$ are the angular coordinates of the extremities: $\theta \in [0, \pi)$, $(\phi, \psi) \in [0, 2\pi)^2$, $\rho_0 \in [-\rho_d, \rho_d]$, $\rho_1 \in [\rho_0, \rho_d]$. The considered cumulative space $\mathcal{B}$ has three dimensions, namely the strip angle axis $\theta$, the distance axis $\rho$ representing $\rho_0$ and $\rho_1$ and the extremity angular coordinate axis $\phi'$ representing $\phi$ and $\psi$. $J_\mathcal{B}$ is the half-line Hough transform (e.g. a *true* pixel votes only for the line segments containing it having the starting point with the lower angular coordinate), and $\mathbb{J}_\mathcal{B}$ is the cumulative space containing $J_\mathcal{B}$ partial sums, derived from Eq. (4) with $l = 5$, $m = 2$, $\alpha_{l-3} = \rho$ and $\alpha_{l-2} = \phi'$.

### 2.2.2 Algorithm

Algorithm 1 describes the way the most significant patterns are detected using the NFA criterion coupled with cumulative spaces. Its inputs are: $I$ the considered *binary* image, the cumulative space $\mathcal{C}$ determined by the considered pattern (for conciseness, $\mathcal{C}$ bi-parameter axes are denoted as 'bip-axes'), and the set of possible patterns $\mathcal{A}$, which is determined by image dimensions and possibly by some application specific constraints. The output of Algorithm 1 is the set of the most significant patterns, $\mathcal{P}$. After the initialization step, the algorithm begins a loop that detects successively the patterns that will be added, one by one, to $\mathcal{P}$ as the most significant pattern at the current iteration. At each iteration, $\mathbb{J}_\mathcal{C}$ is computed according to Eq. (3) or to Eq. (4) depending on the value of $m$ (in this work we focus on $m \in \{1, 2\}$ but the generalization is trivial). Then, two vectors $\boldsymbol{\kappa}[.]$ and $\boldsymbol{\alpha}[.]$ of dimensionality $N$, the number of pixels assumed to be the maximum size of a pattern, are allocated. $\boldsymbol{\kappa}[.]$ elements are integer values and $\boldsymbol{\alpha}[.]$ elements are $l$-tuples. They will store, for each pattern size $j$ in pixel unit, the maximum number of *true* pixels (in $\boldsymbol{\kappa}[j]$) and the corresponding pattern parameter tuple (in $\boldsymbol{\alpha}[j]$). Indeed, for a given pattern area, the significance increases with the number of *true* pixels. Then, it is not necessary to compute the significance values for each pattern, but only for the patterns having different areas (in pixels) and achieving the maximum number of *true* pixels $\boldsymbol{\kappa}[j]$. Thus, the following *for* loop allows for the comparison of the significance of patterns having different pixel sizes. Finally, having found the most significant pattern $\hat{\alpha}$ at the current iteration, we add it to the set of significant patterns $\mathcal{P}$ only if the global significance of the set of patterns increases. If it is not the case, the algorithm ends. Otherwise, before reiteration, the located pattern is removed from the image. In our case, when we remove a pattern, we do not set to *false* all its pixels, but only the exceeding ones relative to naive model parameter $p$. Although sub-optimal, this simple adjustment allows us to avoid penalizing too much patterns which overlap other patterns previously detected.

Note that because of the non monotonicity of the projection of extremities to the strip versus the angular coordinate, in the case of bounded strips,

Eq. (6) should be adapted as follows. If there is an extremum between $\phi$ and $\psi$, consider as new bound the pre-image of the image of $\psi$ closest to $\phi$ (thus enforcing the monotonicity of the projection).

---

**Data**: Binary data image $I$, cumulative space $\mathcal{C}$ having $m$ bip-axes, set of possible
        patterns $\mathcal{A}$;
**Result**: Set of most significant patterns $\mathcal{P}$;
$N \leftarrow$ number of pixels in $I$; $p \leftarrow \frac{\text{number of } true \text{ pixels in } I}{N}$;
Initialize set $\mathcal{P}$ to $\emptyset$ and real number $S_{\mathcal{P}}$ to $0$;
**repeat**
    Initialize $\mathbb{J}_{\mathcal{C}}$ with the transform of $I$ in $\mathcal{C}$;
    Modify $\mathbb{J}_{\mathcal{C}}$ by computing the partial sums along bip-axes;
    Initialize vectors $\boldsymbol{\kappa}\,[.]$, $\boldsymbol{\beta}\,[.]$ of dimensionality $N$ to $\mathbf{0}$;
    **for** *every l-tuple $\boldsymbol{\beta} = (\beta_1, \ldots, \beta_l)$ of $\mathcal{A}$* **do**
        $\mathbf{b} \leftarrow$ pattern of parameters $\boldsymbol{\beta}$;
        $\nu \leftarrow$ area (in pixels) of $\mathbf{b}$ pattern;
        $\kappa\,(\mathbf{b}) \leftarrow$ number of $true$ pixels in $\mathbf{b}$ deduced from points of $\mathbb{J}_{\mathcal{C}}$ (according to
        Eq. (5) if $m = 1$ or Eq. (6) if $m = 2$);
        **if** $\kappa\,(\mathbf{b}) > \boldsymbol{\kappa}\,[\nu]$ **then**
          $\mid$  $\boldsymbol{\kappa}\,[\nu] \leftarrow \kappa\,(\mathbf{b})$; $\boldsymbol{\beta}\,[\nu] \leftarrow \boldsymbol{\beta}$;
        **end**
    **end**
    Initialize $S_{max}$ and $\mathbf{b}$ to $0$;
    **for** $\nu \in [1, N]$ **do**
        **if** $\frac{\boldsymbol{\kappa}\,[\nu]}{\nu} > p$ **then**
          $S \leftarrow S\,(\boldsymbol{\kappa}\,[\nu], \nu, p)$ computed according to Eq. (2);
          **if** $S > S_{max}$ **then**
            $\mid$  $S_{max} \leftarrow S$; $\mathbf{b} \leftarrow$ pattern of parameters $\boldsymbol{\beta}\,[\nu]$;
          **end**
        **end**
    **end**
    $S_{\cup} \leftarrow$ significance of $\mathcal{P} \cup \{\mathbf{b}\}$ computed according to Eq. (2);
    **if** $S_{\cup} > S_{\mathcal{P}}$ **then**
        $\mid$  $\mathcal{P} \leftarrow \mathcal{P} \cup \{\mathbf{b}\}$; $S_{\mathcal{P}} \leftarrow S_{\cup}$; Remove pattern $\mathbf{b}$ from $I$;
    **else**
        $\mid$  $finished = true$
    **end**
**until** *not finished*;

**Algorithm 1:** Detection of the most significant patterns.

Figure 1 illustrates the detection of the parametric patterns introduced in the section. For each case (tile, strip, ring and bounded strip), the first row shows the simulated binary image ($true$ pixels in black and $false$ ones in white) on which the detected pattern frontiers are superimposed in thin light grey. The second row shows the highest significance values versus the cardinality of the considered pattern ($\nu$ parameter in Eq. (1)-(2) and in Algorithm 1). In these figures, each point represents the highest significance value achieved setting the cardinality value $\nu$ and varying the pattern parameters. It means that, given the type of pattern (tile etc.) and the considered binary image, among all the patterns having the same number of pixels ($\nu$ value on x-axis value), the significance achieves its maximum value at the y-value of the point.

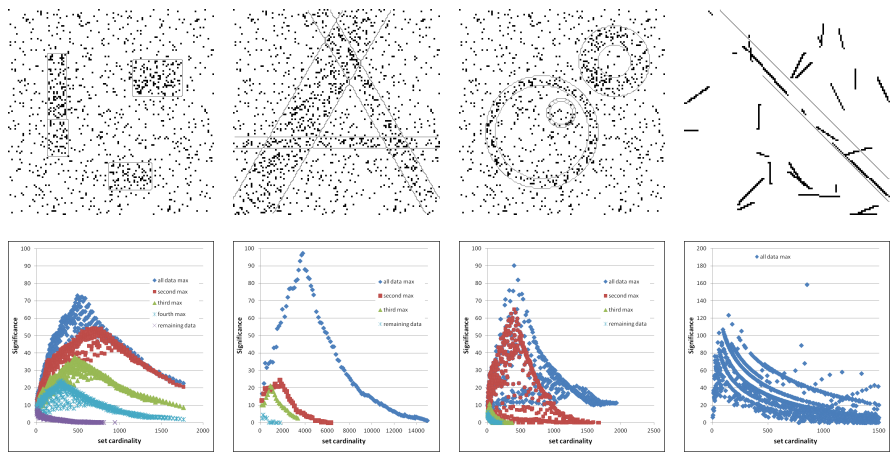

Fig. 1: Examples of parametrized objects detection using cumulative space to compute significance criterion.

In the first three examples, there are three patterns to detect and only one in the last example (bounded strips). The different colors correspond to the different iterations showing the effect of removing a detected pattern from the image. In other words, they show the highest significance points obtained considering either the whole initial set of *true* pixels (presented on first row) or *true* pixel subsets derived by removal of the points belonging to the patterns already detected. Note that in the case of the tile, one of the the patterns is detected in two parts. However, we observe the very good robustness of the proposed detection process. Let us now consider actual data and a real application.

## 3 Crack detection in still images

The proposed approach is suited for applications involving a significance measure or NFA criterion, and benefiting from a finer sampling of the pattern space. For instance, for detection tasks that are quite standard a-contrario problems, the basic idea is that the detection relies on pattern significance (or NFA) that itself depends on the number of *true* pixels belonging to the considered pattern. For instance, for detection estimated at region-level, i.e. in rectangular windows, the refinement of the space would imply that the sampling be performed with a 1-pixel sliding step in both dimensions, rather than using a non-overlapping or half-overlapping window sampling strategy [13]. Undoubtedly, the benefit of the proposed method will be more important for pattern with a high number of parameters, i.e. involving high dimensionality of the parametric space. In this study, we have chosen to illustrate our refinement algorithm on the problem of crack detection.

Crack detection has critical importance for ensuring the security of infrastructures and for minimizing maintenance costs. In terms of appearance, a crack is a discontinuity in the background (i.e. asphalt for roads, concrete for walls or more generally underlying material). Then, it may be detected based on some photometric and geometric features [9] and indeed several proposed approaches (e.g. [31,14]) perform rather well on cracks observed on smooth and homogeneous surfaces (e.g. concrete). However, these methods often fail when the background exhibits noisy texture like in the case of road pavement.

In this study, we focus on noisy background case (including texture, artifacts) and on the dataset proposed in [33]. Since the road surface observations have been acquired by a camera embedded on a vehicle, the road texture induces small clusters of shadow pixels appearing like very numerous dark structures whose density and size vary in the image due to the tilt of the camera (cf. Fig. 2, first column). Therefore, the radiometric features alone do not allow us to distinguish the cracks from the road background and the geometric features should be also considered like in [27]. However, whereas [27] focuses on the modeling of the spatial interactions between line segments, in this work, we focus on the detection of these line segments based on significance (or NFA) computation. The background heterogeneity due to asphalt textons indeed resembles well the null hypothesis. On such a background, the lines that compose the cracks may be seen as a deviation from the naive model, having thus high values of significance.

### 3.1 Related work

Crack detection approaches generally involve a preprocessing step that aims at computing a new data image on which the detection will be easier. The preprocessing step may consist in removing adverse or clutter features (e.g. shadow removal [33]), or in enhancing the pattern e.g. by subtracting the median filtered image [14], or in stressing the filiform feature of the cracks e.g. by using Laplacian of Gaussian or steerable filters [6,16]. In this work, we consider the same preprocessing step as in [3]. Basically, it involves the shadow removal by background subtraction and the estimation of a new radiometric image, whose values gather both gray level information and gradient orientation features.

Then, from the preprocessed image, two analysis scales may be considered for crack detection itself. In [31,20], the detection is based on a local analysis performed across the whole image space using a sliding window, whereas in [33,24,5], the detection relies on a global process related to the expected photometric properties of cracks, based on the computation of minimum cost paths. The local or global search strategy and the parameters required by the algorithms set the scale for the patterns to be detected. Now, we observed that the cracks may be highly variable in terms of scale, thinness and relative contrast. Thus, both local and global scales seem relevant and will then be considered in the proposed approach, through the measure of significance relatively to the context in a multi-scale reasoning.

Finally, note that since the used NFA criterion applies to binary images, we perform an automatic thresholding operation on the pre-processed image to derive the *seed* image, i.e. the binary image where *true* pixels are very likely to belong to the crack. Then, the objective of the whole algorithm presented in next section is to remove the false positives and to correct the false negatives on this *seed* image.

## 3.2 Crack detection algorithm

Following a multi-scale strategy, we adopt a two-step algorithm. The input image is the binary image of the seeds in which the *true* pixels represent (in an incomplete way and including some false positives) the researched patterns. The first step deals with local scale, and aims at determining the most significant local alignments of *true* pixels, that are called *elementary strips* in the following. Then, the second step is intended to identify the significant straight chains of elementary strips.

Algorithm 2 describes the method based on the two successive steps. In Algorithm 2, a window refers to the rectangular image sub-area used for local detection. Its dimensions in columns and rows are given as input parameters. Then, in every considered window, the elementary strips are detected as the most significant unbounded strip(s), following Section 2.2. At the end of this step, a new binary image is derived such that the *true* pixels are exclusively located in detected significant local strips. In other words, the maximization of significance at local scale (over each window) is used as a filtering process which removes a part of the false positives present in the seed image. Besides, the extremities of local strips are also stored as possible extremities of the bounded strips which are estimated in the next step. Then, at image scale, the most significant bounded strip(s) are detected following Section 2.2. Finally, the cracks are approximated by the concatenation of the elementary strips (detected at local scale) that also belong to a bounded strip detected at image scale.

## 3.3 Results

We have applied the proposed algorithm to the public CrackTree dataset provided by [33], which illustrates our approach very well since noisy texture and other degradation artifacts are commonly present on the asphalt surface.

To evaluate quantitatively the obtained results, the precision and recall parameters are computed while distinguishing between the mis-detection and the mis-location of a crack as follows. The true positives and the false positives are computed by comparing the detection results with incrementally dilated ground-truth, while the false negatives are computed comparing the ground-truth with an incrementally dilated version of the detection results. In both cases, the dilation radius corresponds to spatial imprecision.

**Data**: Binary image: $I$; window size;
**Result**: Binary image $I_S$ of cracks approximated by window strips;
$N_c \leftarrow$ number of $I$ columns; $N_r \leftarrow$ number of $I$ rows;
$I' \leftarrow$ binary image of size $N_c \times N_r$ initialized with *false* values;
Set of possible extremities for bounded strips: $\mathcal{E} \leftarrow \emptyset$;
**for** *every non-overlapping window $W$ of $I$* **do**
  $I_W \leftarrow$ the restriction of $I$ to $W$;
  $\mathcal{S}_W \leftarrow$ output of Algorithm 1 having inputs: $I_W$, Hough space having 1 bip axis and $\mathcal{A}$ the set of any strip in $I_W$;
  **for** *every strip $s_k$ of $\mathcal{S}_W$* **do**
    Compute its restriction $s'_{W,k}$ to $W$ in $I$;
    Set the pixels of $s'_{W,k}$ to value *true* in $I'$;
    Add extremities of $s'_{W,k}$ to $\mathcal{E}$;
  **end**
**end**
$\mathcal{S}_{I'} \leftarrow$ output of Algorithm 1 having inputs: $I' \wedge I$, cumulative space for bounded strips having 2 bip axes and $\mathcal{A}$ the set of bounded strips having extremities in $\mathcal{E}$;
$I_S \leftarrow$ binary image of size $N_c \times N_r$ initialized with *false* values;
**for** *every bounded strip $b_k$ of $\mathcal{S}_{I'}$* **do**
  Set the pixels of $b_k$ to value *true* in $I_S$ ;
**end**
$I_S \leftarrow I_S \wedge I'$;

**Algorithm 2**: Crack estimation; $I$ is the image of the seeds, operator $\wedge$ between two binary images is the binary AND operator applied at pixel level.

| dilation radius | 1 | 2 | 3 |
|---|---|---|---|
| (precision, recall) mean values | (86.8, 86.4) | (89.5, 88.7) | (90.7, 89.8) |
| (precision, recall) median values | (87.8, 86.3) | (90.4, 88.9) | (91.6, 89.7) |
| (precision, recall) 75th percentile | (90.6, 91.2) | (93.4, 93.3) | (95.0, 93.9) |
| (precision, recall) 25th percentile | (82.5, 82.3) | (85.8, 85.1) | (86.6, 86.9) |

Table 1: Main statistical values on precision and recall indexes achieved on the CrackTree dataset [33].

Table 1 presents the main statistical values on precision and recall indexes for the considered dataset and increasing dilation radius. For comparison, in [33], the mean precision-recall values were equal to (0.79, 0.92) (accepting an imprecision of 2 pixels) as stated in [33]. We note that precision is improved due to the fact that the analysis at two successive scales allows us to filter most of the false positives.

Now, to provide a deeper and more specific analysis of the obtained results, we have selected some typical examples of results that are shown in the next figures. In these figures, the first column shows the original images; the second column shows in the blue channel the seed image obtained automatically following [3], whereas the results of the local (window based) strip detection appear in red with the window grid in green. The third column shows the final strips (belonging to a significant alignment at global scale) in red overlaid with ground truth in blue. In the last column we show the results of a simple post-processing step which connects the true pixels validated by the final strips using minimum cost paths and possibly remove the obtained path based on its average cost value. Note that this post-processing is not the object of our study but it was necessary to extract the final cracks (for instance for quantitative evaluation).

Figure 2 illustrates how the proposed approach is able to cope with cracks of different width and depth (making the crack more or less dark) and with background texture.

Due to the camera tilt, the background textons are much more prominent in the lower part of the image(cf. $1^{st}$ line example for instance) creating numerous false alarms at pixel level (blue pixels in the images of the second column of Fig. 2). However, local analysis counters the non stationarity of such a noise (caused by textons) by selecting the most significant *elementary strips* within local windows (red segments in the images of the second column of Fig. 2). However, most of the *elementary strips* are false alarms at global scale. Then, global analysis allows for their filtering by checking their consistency *elementary strips* at image scale (red segments in the third column versus the second column in Fig. 2). Finally, having detected the rough shape of the cracks, the post-processing step allows for a finer estimation of the shape as well as removal of some isolated false alarms. Note also that the local analysis is important not only to improve the resilience to texture non stationarity, but also to crack width variations. For instance, the second row illustrates the ability to recover not only main cracks but also thinner ones: the window scale step allows for local significance maximization whereas a global measure would estimate that thin cracks are insignificant relatively to wider cracks.

On Figure 3, examples have been chosen to illustrate the limits of the proposed approach. Three main phenomena induce adverse conditions for cracks detection: the shortness of some crack subparts, the thinness of the crack, and an apparent partial occlusion of the crack. The first phenomenon is illustrated for instance on the two first rows of Figure 3, where some branches of a main crack are missed because they do not present a length sufficient to be considered as significant. The second phenomenon is illustrated on the two last rows of Figure 3, where the thinness of some crack subparts makes them almost invisible. However, in these examples (as in many others but not always), the post-processing allows for the reconnection of the actually detected crack subparts. The third phenomenon is illustrated on the third and the fifth rows of Figure 3, where in some places the crack has been partially filled by the background material inducing a kind of occlusion of the crack. Then, like in the case of the crack extreme thinness, some crack parts are missing at the end of the global scale analysis. Note also the fifth row case is particularly adverse since it combines both thinness and partial occlusion.

Figure 4 presents results on images with a shadow introducing an adverse effect which increases from first to last rows. Indeed, in the $1^{st}$ and $2^{nd}$ line cases, the presence of the shadow only affects the global distribution of gray level values which, according to the shown results, does not actually impact the algorithm performance. The result shown on $3^{rd}$ row allows us to check the robustness to illumination non stationarity induced by the shadow, whereas the result shown on $4^{th}$ row stresses the limit of this robustness. Indeed, when in $3^{rd}$ row case, the local analysis is still able to detect the crack in the shadowed part (because local contrast remains even if attenuated), in $4^{th}$ row case, the darkness of the shadow removes some subparts of the crack that cannot be recovered based on alignment criterion because of the "alligator-skin" feature of the crack.

In summary,we evaluate our algorithm with respect to five adverse phenomena for crack detection: two phenomena that are independent of the cracks themselves, namely the background texture and the shadow presence, and three phenomena that characterize the cracks, namely the thinness, the length of secondary branch(es), and the partial filling (behaving like partial occlusion). The analysis of the results shows that, even in presence of these adverse phenomenons, the algorithm is able to detect almost every strip that composes the actual cracks, and follow their jagged behavior. The window level detection allows us to remove most false alarms introduced at pixel level (seeds), but introduces some window level false alarms (on average, one detection per window is not related to an actual crack) which are then removed by the image scale detection, unless they are found to belong to a significant strip at image level.

## 4 Conclusion

In this paper, a generic method for the NFA criterion computation for pattern detection is proposed. We consider that relying on an advantageous grouping of parameters in the engendered cumulative space is applicable to a wide variety of problems, and may facilitate the use of a-contrario based algorithms for various applications involving parametric pattern detection. Our technique was applied to a crack detection task, which naturally fits the problem as cracks can be seen as a deviation from the naive model in a heterogeneous background, allowing us to illustrate the pertinence and the benefit of re-parametrization and multi-scale a-contrario analysis for complex patterns.

Future work will be devoted to accelerating the accumulation tasks since most cumulative space operations are inherently independent, and they can take advantage of parallel architectures. On such architectures (GPU, FPGA), the available memory resources are often more constrained than on generic systems. However, the reduced memory footprint resulting from our algorithm should be directly beneficial in such scenario.

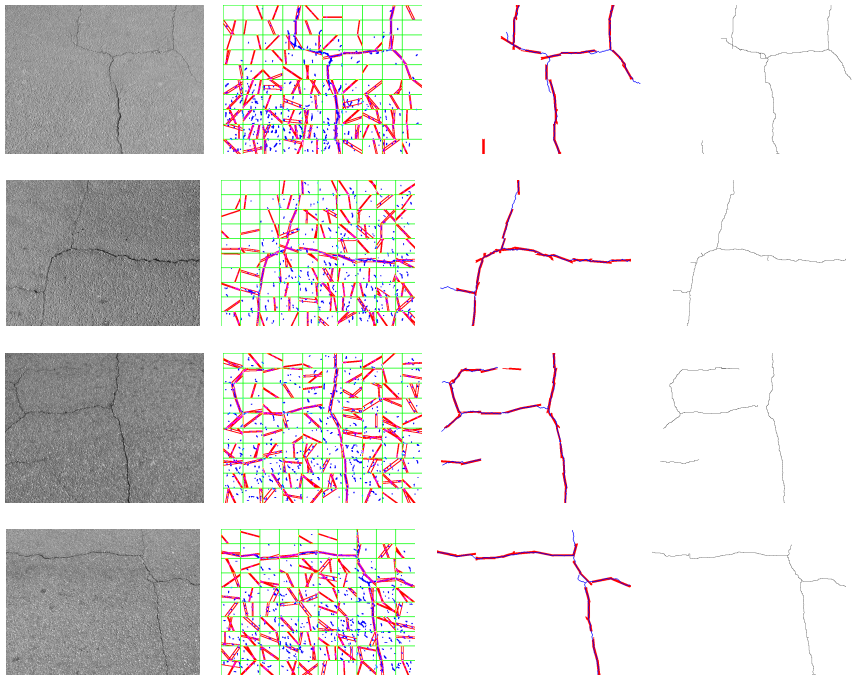

Fig. 2: First set of examples of crack detection, which shows the efficiency of the multi-scale approach for these cracks presenting different width and depth values, as well as background texture: original image ($1^{st}$ column), seeds and results of window level detection ($2^{nd}$ column), comparison between ground truth and image level detection ($3^{rd}$ column), fine crack detection after post-processing ($4^{th}$ column).

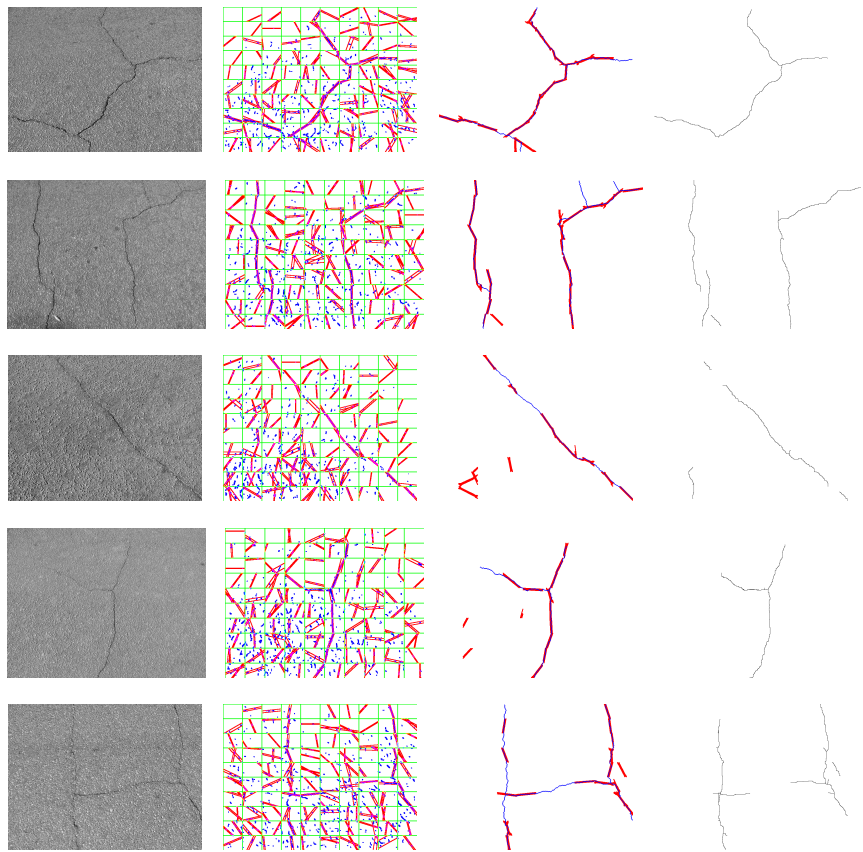

Fig. 3: Second set of examples of crack detection which pushes the boundaries of the proposed approach with respect to either sub-crack shortness, or crack thinness or occlusion of crack subparts: original image ($1^{st}$ column), seeds and results of window level detection ($2^{nd}$ column), comparison between ground truth and image level detection ($3^{rd}$ column), fine crack detection after post-processing ($4^{th}$ column).

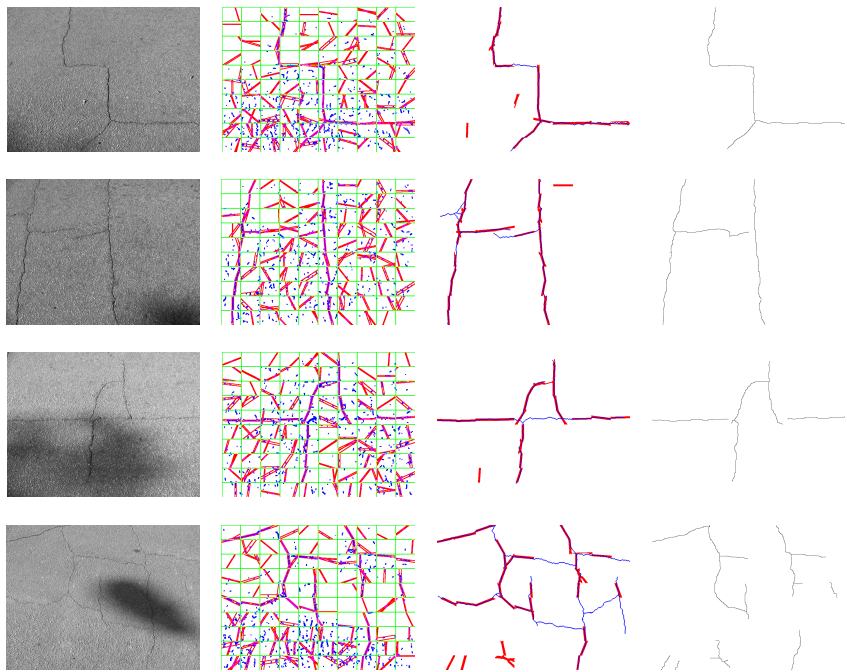

Fig. 4: Third set of examples of crack detection, which underlines the proposed approach robustness with respect to shadows: original image ($1^{st}$ column), seeds and results of window level detection ($2^{nd}$ column), comparison between ground truth and image level detection ($3^{rd}$ column), fine crack detection after post-processing ($4^{th}$ column).